\documentclass{article}

\usepackage[toc,page]{appendix}
\usepackage{tabularx}

\usepackage{graphicx} 
\usepackage{subfigure} 

\usepackage{natbib}

\usepackage[bottom]{footmisc}

\usepackage{amsmath}

\usepackage{array}
\newcolumntype{L}[1]{>{\raggedright\let\newline\\\arraybackslash\hspace{0pt}}m{#1}}
\newcolumntype{C}[1]{>{\centering\let\newline\\\arraybackslash\hspace{0pt}}m{#1}}
\newcolumntype{R}[1]{>{\raggedleft\let\newline\\\arraybackslash\hspace{0pt}}m{#1}}

\usepackage{algorithm}
\usepackage{algorithmic}

\usepackage{siunitx}
\newcommand*\rot{\rotatebox{90}}

\usepackage{hyperref}

\usepackage[accepted]{icml2017}

\icmltitlerunning{Neural Audio Synthesis of Musical Notes with WaveNet Autoencoders}

\title{Neural Audio Synthesis of Musical Notes with WaveNet Autoencoders}

\begin{document} 

\twocolumn[
\icmltitle{Neural Audio Synthesis of Musical Notes \\ with WaveNet Autoencoders}

\icmlsetsymbol{equal}{*}

\begin{icmlauthorlist}
\icmlauthor{Jesse Engel*}{Brain}
\icmlauthor{Cinjon Resnick*}{Brain}
\icmlauthor{Adam Roberts}{Brain}
\icmlauthor{Sander Dieleman}{DeepMind}
\icmlauthor{Douglas Eck}{Brain}
\icmlauthor{Karen Simonyan}{DeepMind}
\icmlauthor{Mohammad Norouzi}{Brain}
\end{icmlauthorlist}

\icmlaffiliation{Brain}{Google Brain, Mountain View, California, USA. Work done while Cinjon Resnick was a mem\-ber of the Google Brain Residency Program}
\icmlaffiliation{DeepMind}{DeepMind, London, England}

\icmlcorrespondingauthor{Jesse Engel}{jesseengel@google.com}

\icmlkeywords{music, generative, nsynth, spectrogram, WaveNet, machine learning, ICML}

\vskip 0.3in

]

\printAffiliationsAndNotice{\icmlEqualContribution}

\begin{abstract} 
Generative models in vision have seen rapid progress due to algorithmic improvements and the availability of high-quality image datasets. 
In this paper, we offer contributions in both these areas to enable similar progress in audio modeling.
First, we detail a powerful new WaveNet-style autoencoder model that conditions an autoregressive decoder on temporal codes learned from the raw audio waveform.
Second, we introduce NSynth, a large-scale and high-quality dataset of musical notes that is an order of magnitude larger than comparable public datasets. 
Using NSynth, we demonstrate improved qualitative and quantitative performance of the WaveNet autoencoder over a well-tuned spectral autoencoder baseline.
Finally, we show that the model learns a manifold of embeddings that allows for morphing between instruments, meaningfully interpolating in timbre to create new types of sounds that are realistic and expressive.

\end{abstract}

\section{Introduction}
\label{Introduction}

 \begin{figure*}[t]
    \includegraphics[width=\textwidth]{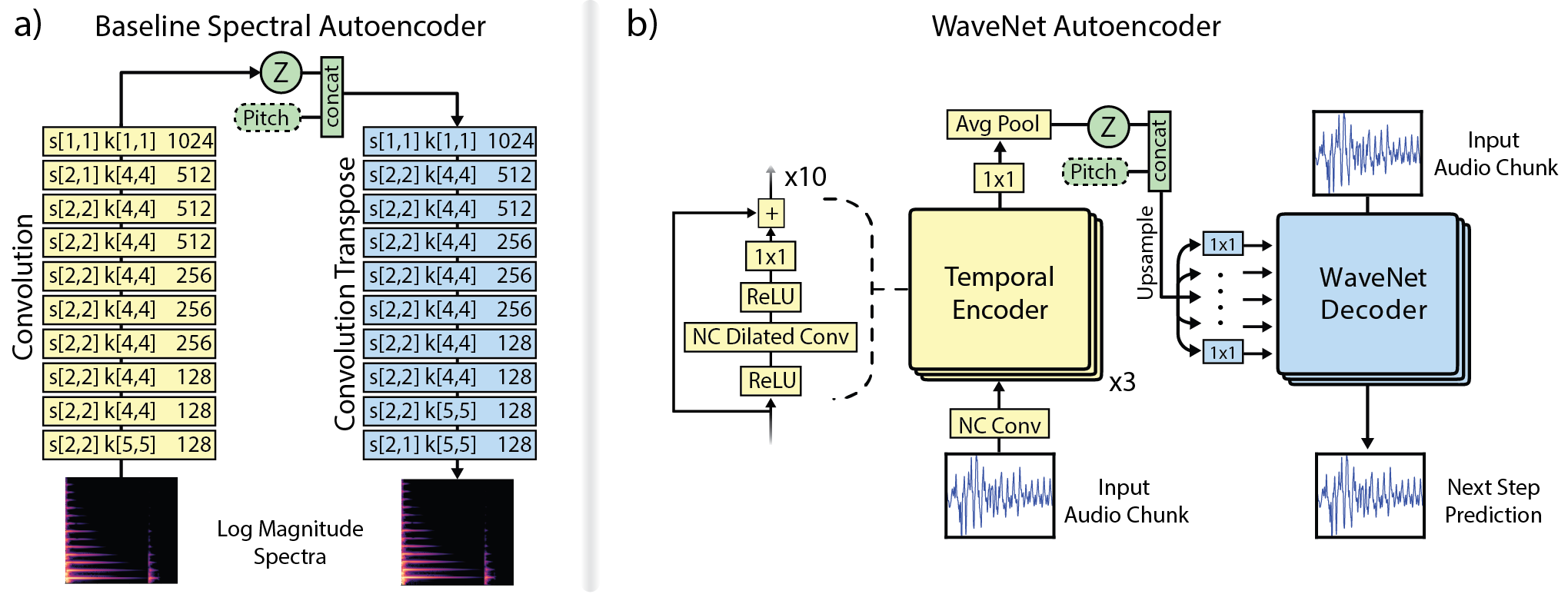}
    \caption{ Models considered in this paper. For both models, we optionally condition on pitch by concatenating the hidden embedding with a one-hot pitch representation. $1a.$ Baseline spectral autoencoder: Each block represents a nonlinear 2-D convolution with stride ($s$), kernel size ($k$), and channels (\#). $1b.$ The WaveNet autoencoder: Downsampling in the encoder occurs only in the average pooling layer. The embeddings are distributed in time and upsampled with nearest neighbor interpolation to the original resolution before biasing each layer of the decoder. `NC' indicates non-causal convolution. `1x1' indicates a 1-D convolution with kernel size 1. See Section~\ref{sec:WaveNetAutoencoder} for further details.}
    \label{fig:WN-AE}
\end{figure*}

Audio synthesis is important for a large range of applications including text-to-speech (TTS) systems and music generation. Audio generation algorithms, know as vocoders in TTS and synthesizers in music, respond to higher-level control signals to create fine-grained audio waveforms. Synthesizers have a long history of being hand-designed instruments, accepting control signals such as `pitch', `velocity', and filter parameters to shape the tone, timbre, and dynamics of a sound \cite{pinch2009analog}. In spite of their limitations, or perhaps because of them, synthesizers have had a profound effect on the course of music and culture in the past half century \cite{DP}. 

In this paper, we outline a data-driven approach to audio synthesis. Rather than specifying a specific arrangement of oscillators or an algorithm for sample playback, such as in FM Synthesis or Granular Synthesis \cite{chowning1973synthesis, xenakis1971formalized}, we show that it is possible to generate new types of expressive and realistic instrument sounds with a neural network model. Further, we show that this model can learn a semantically meaningful hidden representation that can be used as a high-level control signal for manipulating tone, timbre, and dynamics during playback. 

Explicitly, our two contributions to advance the state of generative audio modeling are:

\begin{itemize}
\item A WaveNet-style autoencoder that learns temporal hidden codes to effectively capture longer term structure without external conditioning.
\item NSynth: a large-scale dataset for exploring neural audio synthesis of musical notes. 
\end{itemize}

The primary motivation for our novel autoencoder structure follows from the recent advances in autoregressive models like WaveNet \cite{DBLP:journals/corr/OordDZSVGKSK16} and SampleRNN \cite{DBLP:journals/corr/MehriKGKJSCB16}. They have proven to be effective at modeling short and medium scale ($\sim$500ms) signals, but rely on external conditioning for longer-term dependencies. Our autoencoder removes the need for that external conditioning. It consists of a WaveNet-like encoder that infers hidden embeddings distributed in time and a WaveNet decoder that uses those embeddings to effectively reconstruct the original audio. This structure allows the size of an embedding to scale with the size of the input and encode over much longer time scales.

Recent breakthroughs in generative modeling of images \cite{kingma2013auto, goodfellow2014generative, DBLP:journals/corr/OordKK16} have been predicated on the availability of high-quality and large-scale datasets such as MNIST \cite{lecun1998mnist}, SVHN \cite{netzer2011reading}, CIFAR \cite{krizhevsky2009learning} and ImageNet \cite{imagenet_cvpr09}. While generative models are notoriously hard to evaluate \cite{theis2015note}, these datasets provide a common test bed for consistent qualitative and quantitative evaluation, such as with the use of the Inception score \cite{DBLP:journals/corr/SalimansGZCRC16}.

We recognized the need for an audio dataset that was as approachable as those in the image domain. Audio signals found in the wild contain multi-scale dependencies that prove particularly difficult to model \cite{raffel2016learning, bertin2011million, king2008blizzard, thickstun2016learning}, leading many previous efforts at data-driven audio synthesis to focus on more constrained domains such as texture synthesis or training small parametric models \cite{sarroff2014musical, mcdermott2009sound}. 

Inspired by the large, high-quality image datasets, NSynth is an order of magnitude larger than comparable public datasets \cite{MINST}. It consists of $\sim$300k four-second annotated notes sampled at 16kHz from $\sim$1k harmonic musical instruments. 

After introducing the models and describing the dataset, we evaluate the performance of the WaveNet autoencoder over a baseline convolutional autoencoder model trained on spectrograms. We examine the tasks of reconstruction and interpolation, and analyze the learned space of embeddings. For qualitative evaluation, download audio files for all examples mentioned in this paper \href{https://download.magenta.tensorflow.org/audio_examples/nsynth/nsynth_audio_examples.zip}{\textbf{here}}. Despite our best efforts to convey analysis in plots, \emph{listening to the samples is essential to understanding this paper} and we strongly encourage the reader to listen along as they read.

\section{Models}

\subsection{WaveNet Autoencoder}
\label{sec:WaveNetAutoencoder}

WaveNet \cite{DBLP:journals/corr/OordDZSVGKSK16} is a powerful generative approach to probabilistic modeling of raw audio. In this section we describe our novel WaveNet autoencoder structure. The primary motivation for this approach is to attain consistent long-term structure without external conditioning. A secondary motivation is to use the learned encodings for applications such as meaningful audio interpolation.

Recalling the original WaveNet architecture described in \cite{DBLP:journals/corr/OordDZSVGKSK16}, at each step a stack of dilated convolutions predicts the next sample of audio from a fixed-size input of prior sample values. The joint probability of the audio $x$ is factorized as a product of conditional probabilities:

\begin{align*}
p(x) = \prod_{i=1}^N p(x_i | x_1, ..., x_{N-1})
\end{align*}


Unconditional generation from this model manifests as ``babbling" due to the lack of longer term structure (Listen: \emph{CAUTION, VERY LOUD!} (\href{https://download.magenta.tensorflow.org/audio_examples/nsynth/UnconditionalWaveNetGeneration_WARNING_HIGH_VOLUME/0.mp3}{ex1},
\href{https://download.magenta.tensorflow.org/audio_examples/nsynth/UnconditionalWaveNetGeneration_WARNING_HIGH_VOLUME/1.mp3}{ex2},
\href{https://download.magenta.tensorflow.org/audio_examples/nsynth/UnconditionalWaveNetGeneration_WARNING_HIGH_VOLUME/2.mp3}{ex3},
\href{https://download.magenta.tensorflow.org/audio_examples/nsynth/UnconditionalWaveNetGeneration_WARNING_HIGH_VOLUME/3.mp3}{ex4})). However, \cite{DBLP:journals/corr/OordDZSVGKSK16} showed in the context of speech that long-range structure can be enforced by conditioning on temporally aligned linguistic features. 

Our autoencoder removes the need for that external conditioning. It works by taking raw audio waveform as input from which the encoder produces an embedding $Z = f(x)$. Next, we causally shift the same input and feed it into the decoder, which reproduces the input waveform. The joint probablity is now:

\begin{align*}
p(x) = \prod_{i=1}^N p(x_i | x_1, ..., x_{N-1}, f(x)) 
\end{align*}

We could parameterize $Z$ as a latent variable $p(Z | x)$ that we would have to marginalize over \cite{DBLP:journals/corr/GulrajaniKATVVC16}, but in practice we have found this to be less effective. As discussed in \cite{DBLP:journals/corr/ChenKSDDSSA16}, this may be due to the decoder being so powerful that it can ignore the latent variables unless they encode a much larger context that's otherwise inaccessible.

Note that the decoder could completely ignore the deterministic encoding and degenerate to a standard unconditioned WaveNet. However, because the encoding is a strong signal for the supervised output, the model learns to utilize it.

During inference, the decoder autoregressively generates a single output sample at a time conditioned on an embedding and a starting palette of zeros. The embedding can be inferred deterministically from audio or drawn from other points in the embedding space, e.g. through interpolation or analogy \cite{DBLP:journals/corr/White16a}.

Figure \ref{fig:WN-AE}b depicts the model architecture in more detail. The temporal encoder model is a 30-layer nonlinear residual network of dilated convolutions followed by 1x1 convolutions. Each convolution has 128 channels and precedes a ReLU nonlinearity. The output feed into another 1x1 convolution before downsampling with average pooling to get the encoding $Z$. 
We call it a `temporal encoding' because the result is a sequence of hidden codes with separate dimensions for time and channel. The time resolution depends on the stride of the pooling. We tune the stride, keeping total size of the embedding constant ($\sim$32x compression). In the trade-off between temporal resolution and embedding expressivity, we find a sweet spot at a stride of 512 (32ms) with 16 dimensions per timestep, yielding a 125x16 embedding for each NSynth note. We additionally explore models that condition on global attributes by utilizing a one-hot pitch embedding.

The WaveNet decoder model is similar to that presented in \cite{DBLP:journals/corr/OordDZSVGKSK16}. We condition it by biasing every layer with a different linear projection of the temporal embeddings. Since the decoder does not downsample anywhere in the network, we upsample the temporal encodings to the original audio rate with nearest neighbor interpolation. As in the original design, we quantize our input audio using 8-bit mu-law encoding and predict each output step with a softmax over the resulting 256 values.

This WaveNet autoencoder is a deep and expressive network, but has the trade-off of being limited in temporal context to the chunk-size of the training audio. While this is sufficient for consistently encoding the identity of a sound and interpolating among many sounds, achieving larger context would be better and is an area of ongoing research.

\subsection{Baseline: Spectral Autoencoder}
As a point of comparison, we set out to create a straightforward yet strong baseline for the our neural audio synthesis experiments. Inspired by image models \cite{vincent2010stacked}, we explore convolutional autoencoder structures with a bottleneck that forces the model to find a compressed representation for an entire note. Figure~\ref{fig:WN-AE}a shows a block diagram of our baseline architecture. The convolutional encoder and decoder are each 10 layers deep with 2x2 strides and 4x4 kernels. Every layer is followed by a leaky-ReLU (0.1) nonlinearity and batch normalization \cite{DBLP:journals/corr/IoffeS15}. The number of channels grows from 128 to 1024 before a linear fully-connected layer creates a single 1984\footnote{This size was aligned with a WaveNet autoencoder that had a pooling stride of 1024 and a 62x32 embedding.} dimensional hidden vector ($Z$) to match that of the WaveNet autoencoder. 

Given the simplicity of the architecture, we examined a range of input representations. Using the raw waveform as input with a mean-squared error (MSE) cost proved difficult to train and highlighted the inadequacy of the independent Gaussian assumption. Spectral representations such as the real and imaginary components of the Fast Fourier Transform (FFT) fared better, but suffered from low perceptual quality despite achieving low MSE cost. We found that training on the log magnitude of the power spectra, peak normalized to be between 0 and 1, correlated better with perceptual distortion. 

We also explored several representations of phase, including instantaneous frequency and circular normal cost functions (see Appendix), but in each case independently estimating phase and magnitude led to poor sample quality due to phase errors. We find a large improvement by estimating only the magnitude and using a well established iterative technique to reconstruct the phase \cite{griffin1984signal}. To get the best results, we used a large FFT size (1024) relative to the hop size (256) and ran the algorithm for 1000 iterations. As a final heuristic, we weighted the MSE loss, starting at 10 for 0Hz and decreasing linearly to 1 at 4000Hz and above. At the expense of some precision in timbre, this created more phase coherence for the fundamentals of notes, where errors in the linear spectrum lead to a larger relative error in frequency.

\subsection{Training}

We train all models with stochastic gradient descent with an Adam optimizer \cite{DBLP:journals/corr/KingmaB14}. The baseline models commonly use a learning rate of 1e-4, while the WaveNet models use a schedule, starting at 2e-4 and descending to 6e-5, 2e-5, and 6e-6 at iterations 120k, 180k, and 240k respectively. The baseline models train asynchronously for 1800k iterations with a batch size of 8. The WaveNet models train synchronously for 250k iterations with a batch size of 32.

\section{The NSynth Dataset} 

To evaluate our WaveNet autoencoder model, we wanted an audio dataset that let us explore the learned embeddings. Musical notes are an ideal setting for this study as we hypothesize that the embeddings will capture structure such as pitch, dynamics, and timbre. While several smaller datasets currently exist \cite{goto2003rwc, romani2015real}, deep networks train better on abundant, high-quality data, motivating the development of a new dataset. 

\subsection{A Dataset of Musical Notes}
NSynth consists of \num{306043} musical notes, each with a unique pitch, timbre, and envelope. For \num{1006} instruments from commercial sample libraries, we generated four second, monophonic 16kHz audio snippets, referred to as notes, by ranging over every pitch of a standard MIDI piano (21-108) as well as five different velocities\footnote{MIDI velocity is similar to volume control and they have a direct relationship. For physical intuition, higher velocity corresponds to pressing a piano key harder.} (25, 50, 75, 100, 127). The note was held for the first three seconds and allowed to decay for the final second. 
Some instruments are not capable of producing all 88 pitches in this range, resulting in an average of 65.4 pitches per instrument. Furthermore, the commercial sample packs occasionally contain duplicate sounds across multiple velocities, leaving an average of 4.75 unique velocities per pitch. 

\begin{figure*}[t]
    \centering
    \includegraphics[width=\textwidth,height=.63\textwidth]{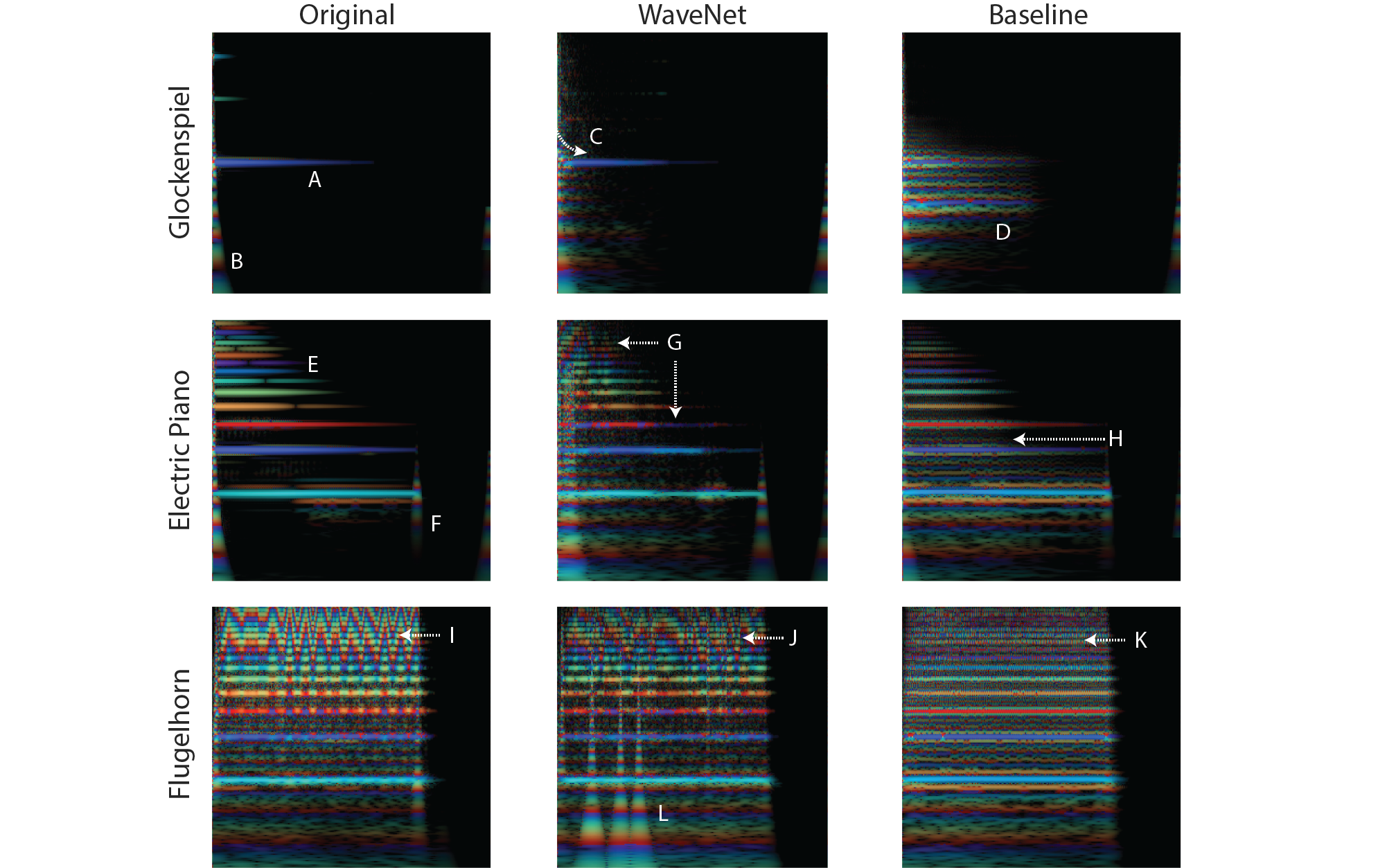}
    \caption{Reconstructions of notes from three different instruments. Each note is displayed as a "Rainbowgram", a CQT spectrogram with intensity of lines proportional to the log magnitude of the power spectrum and color given by the derivative of the phase. Time is on the horizontal axis and frequency on the vertical axis.  See Section~\ref{sec:Reconstruction} for details. (Listen: 
    Glockenspiel (\href{https://download.magenta.tensorflow.org/audio_examples/nsynth/Figure2_Reconstruction/Originals/Glockenspiel.mp3}{O},
    \href{https://download.magenta.tensorflow.org/audio_examples/nsynth/Figure2_Reconstruction/WaveNet/Glockenspiel.mp3}{W},
    \href{https://download.magenta.tensorflow.org/audio_examples/nsynth/Figure2_Reconstruction/Baseline/Glockenspiel.mp3}{B}), 
    Electric Piano (\href{https://download.magenta.tensorflow.org/audio_examples/nsynth/Figure2_Reconstruction/Originals/ElectricPiano.mp3}{O},
    \href{https://download.magenta.tensorflow.org/audio_examples/nsynth/Figure2_Reconstruction/WaveNet/ElectricPiano.mp3}{W},
    \href{https://download.magenta.tensorflow.org/audio_examples/nsynth/Figure2_Reconstruction/Baseline/ElectricPiano.mp3}{B}), 
    Flugelhorn (\href{https://download.magenta.tensorflow.org/audio_examples/nsynth/Figure2_Reconstruction/Originals/Flugelhorn.mp3}{O},
    \href{https://download.magenta.tensorflow.org/audio_examples/nsynth/Figure2_Reconstruction/WaveNet/Flugelhorn.mp3}{W},
    \href{https://download.magenta.tensorflow.org/audio_examples/nsynth/Figure2_Reconstruction/Baseline/Flugelhorn.mp3}{B}))}
    \label{fig:Reconstruction}
\end{figure*}

\subsection{Annotations}
We also annotated each of the notes with three additional pieces of information based on a combination of human evaluation and heuristic algorithms:
\begin{itemize}  
\item Source: The method of sound production for the note's instrument. This can be one of `acoustic' or `electronic' for instruments that were recorded from acoustic or electronic instruments, respectively, or `synthetic' for synthesized instruments.
\item Family: The high-level family of which the note's instrument is a member. Each instrument is a member of exactly one family. See Appendix for the complete list.
\item Qualities: Sonic qualities of the note. See Appendix for the complete list of classes and their co-occurrences. Each note is annotated with zero or more qualities.
\end{itemize}

\subsubsection{Availability}
The full NSynth dataset is available for download at \mbox{\url{https://magenta.tensorflow.org/datasets/nsynth}} as TFRecord files split into training and holdout sets. Each note is represented by a serialized TensorFlow Example protocol buffer containing the note and annotations. Details of the format can be found in the README.

\begin{figure*}[t]
    \centering
    \includegraphics[width=\textwidth,height=.63\textwidth]{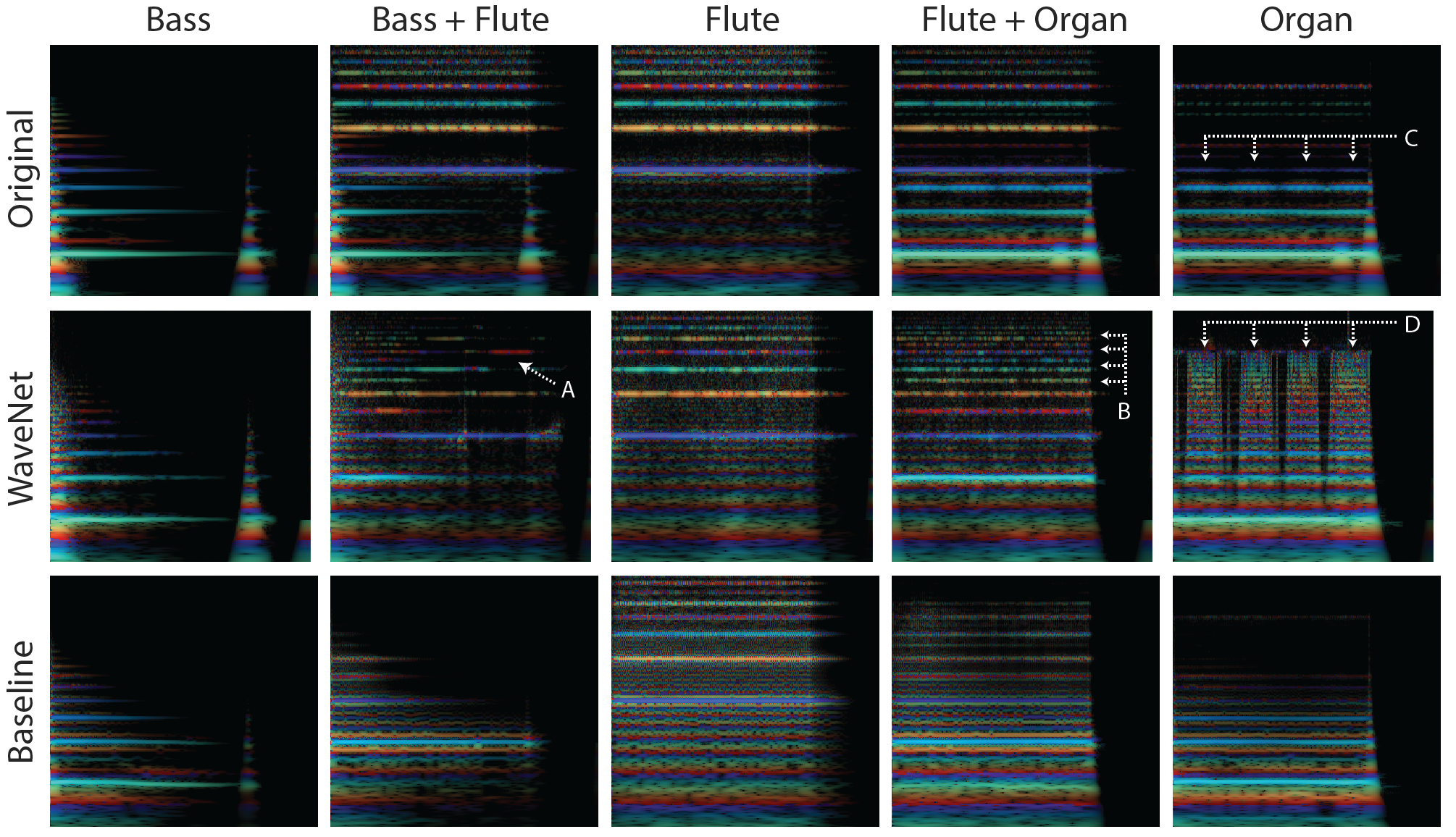}
    \caption{Rainbowgrams of linear interpolations between three different notes from instruments in the holdout set. For the original rainbowgrams, the raw audio is linearly mixed. For the models, samples are generated from linear interpolations in embedding space. See Section~\ref{sec:InterpolateZ} for details.(Listen: 
    Original (\href{https://download.magenta.tensorflow.org/audio_examples/nsynth/Figure3_Interpolation/Original/0_Bass.mp3}{B},
    \href{https://download.magenta.tensorflow.org/audio_examples/nsynth/Figure3_Interpolation/Original/1_Bass+Flute.mp3}{BF},    
    \href{https://download.magenta.tensorflow.org/audio_examples/nsynth/Figure3_Interpolation/Original/2_Flute.mp3}{F},    
    \href{https://download.magenta.tensorflow.org/audio_examples/nsynth/Figure3_Interpolation/Original/3_Flute+Organ.mp3}{FO},    
    \href{https://download.magenta.tensorflow.org/audio_examples/nsynth/Figure3_Interpolation/Original/4_Organ.mp3}{O},    
    \href{https://download.magenta.tensorflow.org/audio_examples/nsynth/Figure3_Interpolation/Original/5_Organ+Bass.mp3}{OB}),    
    WaveNet (\href{https://download.magenta.tensorflow.org/audio_examples/nsynth/Figure3_Interpolation/WaveNet/0_Bass.mp3}{B},
    \href{https://download.magenta.tensorflow.org/audio_examples/nsynth/Figure3_Interpolation/WaveNet/1_Bass+Flute.mp3}{BF},    
    \href{https://download.magenta.tensorflow.org/audio_examples/nsynth/Figure3_Interpolation/WaveNet/2_Flute.mp3}{F},    
    \href{https://download.magenta.tensorflow.org/audio_examples/nsynth/Figure3_Interpolation/WaveNet/3_Flute+Organ.mp3}{FO},    
    \href{https://download.magenta.tensorflow.org/audio_examples/nsynth/Figure3_Interpolation/WaveNet/4_Organ.mp3}{O},    
    \href{https://download.magenta.tensorflow.org/audio_examples/nsynth/Figure3_Interpolation/WaveNet/5_Organ+Bass.mp3}{OB}),    
    Baseline (\href{https://download.magenta.tensorflow.org/audio_examples/nsynth/Figure3_Interpolation/Baseline/0_Bass.mp3}{B},
    \href{https://download.magenta.tensorflow.org/audio_examples/nsynth/Figure3_Interpolation/Baseline/1_Bass+Flute.mp3}{BF},    
    \href{https://download.magenta.tensorflow.org/audio_examples/nsynth/Figure3_Interpolation/Baseline/2_Flute.mp3}{F},    
    \href{https://download.magenta.tensorflow.org/audio_examples/nsynth/Figure3_Interpolation/Baseline/3_Flute+Organ.mp3}{FO},    
    \href{https://download.magenta.tensorflow.org/audio_examples/nsynth/Figure3_Interpolation/Baseline/4_Organ.mp3}{O},    
    \href{https://download.magenta.tensorflow.org/audio_examples/nsynth/Figure3_Interpolation/Baseline/5_Organ+Bass.mp3}{OB}))}    
    \label{fig:InterpolateZ}
\end{figure*}

\section{Evaluation}

We evaluate and analyze our models on the tasks of note reconstruction, instrument interpolation, and pitch interpolation. 

Audio is notoriously hard to represent visually. Magnitude spectrograms capture many aspects of a signal for analytics, but two spectrograms that appear very similar to the eye can correspond to audio that sound drastically different due to phase differences. We have included supplemental audio examples of every plot and \emph{encourage the reader to listen along as they read}.

That said, in our analysis we present examples as plots of the constant-q transform (CQT) \cite{brown1991calculation}, which is useful because it is shift invariant to changes in the fundamental frequency. In this way, the structure and envelope of the overtone series (higher harmonics) determines the dynamics and timbre of a note, regardless of its base frequency. However, due to the logarithmic binning of frequencies, transient noise-like impulses appear as rainbow ``pyramidal spikes" rather than straight broadband lines. We display CQTs with a pitch range of 24-96 (C2-C8), hop size of 256, 40 bins per octave, and a filter scale of 0.8.

As phase plays such an essential part in sample quality, we have attempted to show both magnitude and phase on the same plot. The intensity of lines is proportional to the log magnitude of the power spectrum while the color is given by the derivative of the unrolled phase (`instantaneous frequency') \cite{boashash1992estimating}. We display the derivative of the phase because it creates a solid continuous line for a harmonic of a consistent frequency. We can understand this because if the instantaneous frequency of a harmonic ($f_{harm}$) and an FFT bin ($f_{bin}$) are not exactly equal, each timestep will introduce a constant phase shift, $\Delta\phi = (f_{bin} - f_{harm}) \frac{hop size}{sample rate}$. We affectionately refer to these instantaneous frequency colored spectrograms as "Rainbowgrams" due to their tendency to form rainbows as the instantaneous frequencies modulate up and down.

\subsection{Reconstruction}
\label{sec:Reconstruction}

Figure~\ref{fig:Reconstruction} displays rainbowgrams for notes from 3 different instruments in the holdout set, where the original notes are on the first column and the model reconstructions are on the second and third columns. Each note has a similar structure with some noise on onset, a fundamental frequency with a series of harmonics, and a decay. For all the WaveNet models, there is a slight built-in distortion due to the compression of the mu-law encoding. It is a minor effect for many samples, but is more pronounced for lower frequencies. Using different representations without this distortion is an ongoing area of research. 

While each rainbowgram matches the general contour of the original note, we can hear a pronounced difference in sample quality that we can ascribe to certain features. For the Glockenspiel, we can see that the WaveNet autoencoder reproduces the magnitude and phase of the fundamental (solid blue stripe, (A)), and also the noise on attack (vertical rainbow spike (B)). There is a slight error in the fundamental as it starts a little high and quickly descends to the correct pitch (C). In contrast, the baseline has a more percussive, multitonal sound, similar to a bell or gong. The fundamental is still present, but so are other frequencies, and the phases estimated from the Griffin-Lim procedure are noisy as indicated by the blurred horizontal rainbow texture (D).   

The electric piano has a more clearly defined harmonic series (the horizontal rainbow solid lines, (E)) and a noise on the beginning and end of the note (vertical rainbow spikes, (F)). Listening to the sound, we hear that it is slightly distorted, which promotes these upper harmonics. Both the WaveNet autoencoder and the baseline produce rainbowgrams with similar shapes to the original, but with different types of phase artifacts. The WaveNet model has sufficient phase structure to model the distortion, but has a slight wavering of the instantaneous frequency of some harmonics, as seen in the color change in harmonic stripes (G). In contrast, the baseline lacks the structure in phase to maintain the punchy character of the original note, and produces a duller sound that is slightly out of tune. This is represented in the less brightly colored harmonics due to phase noise (H). 

The flugelhorn displays perhaps the starkest difference between the two models. The sound combines rich harmonics (many lines), non-tonal wind and lip noise (background color), and vibrato - oscillation of pitch that results in a corresponding rainbow of color in all of the harmonics. While the WaveNet autoencoder does not replicate the exact trace of the vibrato (I), it creates a very similar rainbowgram with oscillations in the instantaneous frequency at all levels synced across the harmonics (J). This results in a rich and natural sounding reconstruction with all three aspects of the original sound. The baseline, by comparison, is unable to model such structure. It creates a more or less correct harmonic series, but the phase has lots of random perturbations. Visually this shows up as colors which are faded and speckled with rainbow noise (K), which contrasts with the bright colors of the original and WaveNet examples. Acoustically, this manifests as an unappealing buzzing sound laid over an inexpressive and consistent series of harmonics. The WaveNet model also produces a few inaudible discontinuities visually evidenced by the vertical rainbow spikes (L).

\subsubsection{Quantitative Comparison}
Inspired by the use of the Inception Score for images \cite{DBLP:journals/corr/SalimansGZCRC16}, we train a multi-task classification network to perform a quantitative comparison of the model reconstructions by predicting pitch and quality labels on the NSynth dataset (details in the Appendix). The network configuration is the same as the baseline encoder and testing is done on reconstructions of a randomly chosen subset of 4096 examples from the held-out set. 

\begin{table}[!htbp]
\caption{Classification accuracy of a deep nonlinear pitch and quality classifier on reconstructions of a test set.}
\label{table:deep-classifier}
\begin{center}
\begin{small}
\begin{sc}
\begin{tabular}{l c c}
\abovespace
 & Pitch & Quality \\
\hline\abovespace
Original Audio & 91.6\% & 90.1\% \\
WaveNet Recon & 79.6\% & 88.9\% \\
Baseline Recon & 46.9\% & 85.2\% \\
\hline
\end{tabular}
\end{sc}
\end{small}
\end{center}
\end{table}

\begin{figure*}[t]
    \includegraphics[width=\textwidth]{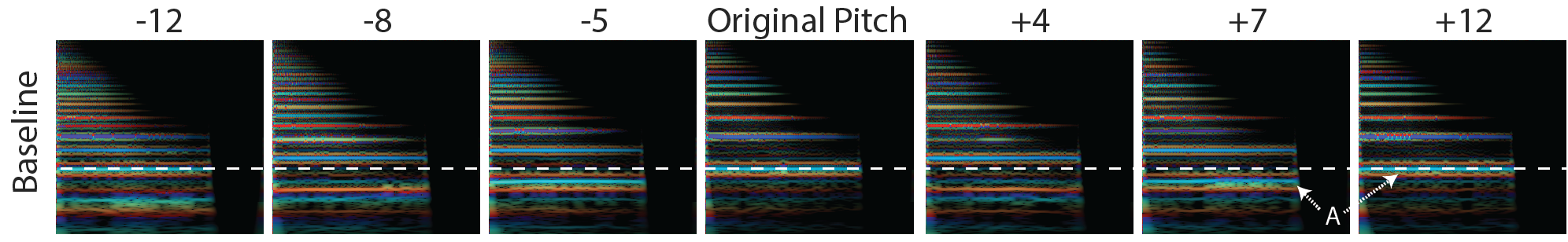}
    \caption{Conditioning on pitch. These rainbowgrams are reconstructions of a single electric piano note from the holdout set. They were synthesized with the baseline model (128 hidden dimensions). By holding $Z$ constant and conditioning on different pitches, we can play two octaves of a C major chord from a single embedding. The original pitch (MIDI C60) is dashed in white for comparison. See Section~\ref{sec:PitchShift} for details. (Listen: \href{https://download.magenta.tensorflow.org/audio_examples/nsynth/Figure4_Pitch/Baseline/0_pitch_-12.mp3}{-12}, \href{https://download.magenta.tensorflow.org/audio_examples/nsynth/Figure4_Pitch/Baseline/1_pitch_-8.mp3}{-8}, \href{https://download.magenta.tensorflow.org/audio_examples/nsynth/Figure4_Pitch/Baseline/2_pitch_-5.mp3}{-5}, \href{https://download.magenta.tensorflow.org/audio_examples/nsynth/Figure4_Pitch/Baseline/3_pitch_0.mp3}{0}, \href{https://download.magenta.tensorflow.org/audio_examples/nsynth/Figure4_Pitch/Baseline/4_pitch_+4.mp3}{+4}, \href{https://download.magenta.tensorflow.org/audio_examples/nsynth/Figure4_Pitch/Baseline/5_pitch_+7.mp3}{+7}, \href{https://download.magenta.tensorflow.org/audio_examples/nsynth/Figure4_Pitch/Baseline/6_pitch_+12.mp3}{+12})}
    \label{fig:PitchShift}
\end{figure*}

\begin{figure}[ht]
\vskip 0.1in
\begin{center}
\centerline{\includegraphics[width=\columnwidth,height=\columnwidth]{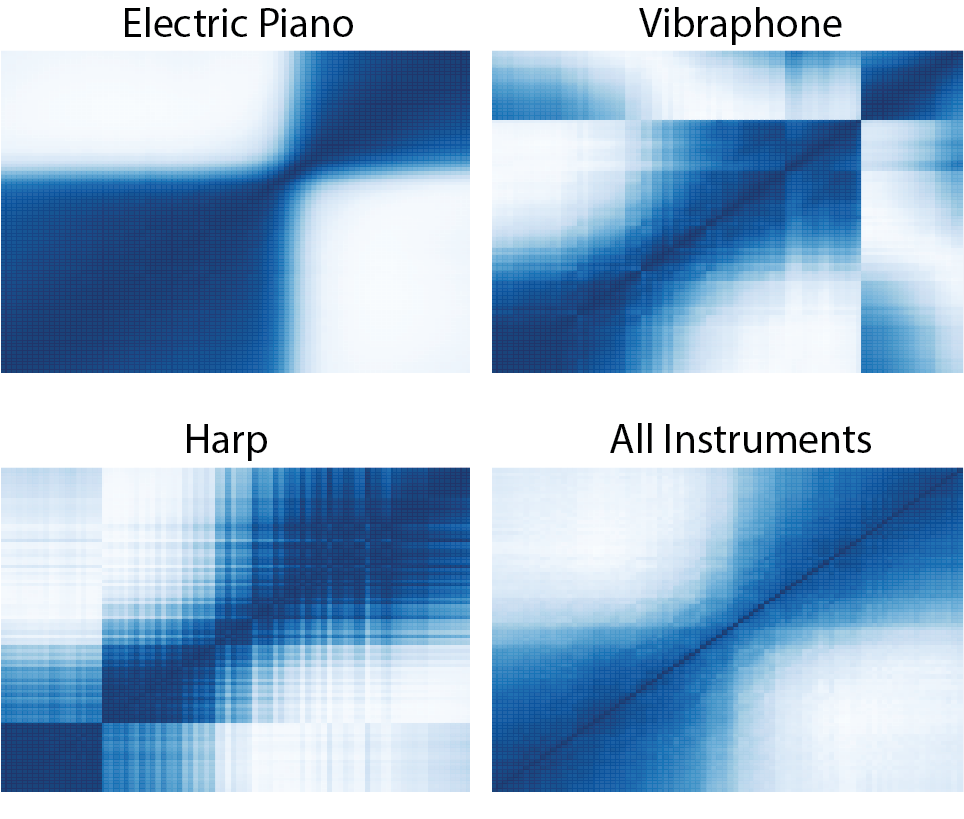}}
\caption{Correlation of embeddings across pitch for three different instruments and the average across all instruments. These embeddings were taken from a WaveNet model trained without pitch conditioning.}
\label{fig:Corr}
\end{center}
\vskip -0.1in
\end{figure}

The results in Table~\ref{table:deep-classifier} confirm our qualititive observation that the WaveNet reconstructions are of superior quality. The classifier is $\sim$70\% more successful at extracting pitch from the reconstructed WaveNet samples than the baseline and several points higher for predicting quality information, giving an accuracy roughly equal to the original audio.

\subsection{Interpolation in Timbre and Dynamics}
\label{sec:InterpolateZ}

Given the limited factors of variation in the dataset, we know that a successful embedding space ($Z$) should span the range of timbre and dynamics in its reconstructions. In Figure~\ref{fig:InterpolateZ}, we show reconstructions from linear interpolations (0.5:0.5) in the $Z$ space among three different instruments and additionally compare these to interpolations in the original audio space. The latter are simple super-positions of the individual instruments' rainbowgrams. This is perceptually equivalent to the two instruments being played at the same time.

In contrast, we find that the generative models fuse aspects of the instruments. As we saw in Section~\ref{sec:Reconstruction}, the WaveNet autoencoder models the data much more realistically than the baseline, so it is no surprise that it also learns a manifold of codes that yield more perceptually interesting reconstructions.

For example, in the interpolated note between the bass and flute (Figure~\ref{fig:InterpolateZ}, column 2), we can hear and see that both the baseline and WaveNet models blend the harmonic structure of the two instruments while imposing the amplitude envelope of the bass note onto the upper harmonics of the flute note. However, the WaveNet model goes beyond this to create a dynamic mixing of the overtones in time, even jumping to a higher harmonic at the end of the note (A). This sound captures expressive aspects of the timbre and dynamics of both the bass and flute, but is distinctly separate from either original note. This contrasts with the interpolation in audio space, where the dynamics and timbre of the two notes is independent. The baseline model also introduces phase distortion similar to those in the reconstructions of the bass and flute.

We see this phenomenon again in the interpolation between flute and organ (Figure~\ref{fig:InterpolateZ}, column 4). Both models also seem to create new harmonic structure, rather than just overlay the original harmonics. The WaveNet model adds additional harmonics as well as a sub-harmonic to the original flute note, all while preserving phase relationships (B). The resulting sound has the breathiness of a flute, with the upper frequency modulation of an organ. By contrast, the lack of phase structure in the baseline leads to a new harmonic yet dull sound lacking a unique character. 

The WaveNet model additionally has a tendency to exaggerate amplitude modulation behavior, while the baseline suppresses it. If we examine the original organ sound (Figure~\ref{fig:InterpolateZ}, column 5), we can see a subtle modulation signified by the blue harmonics periodically fading to black (C). The baseline model misses this behavior completely as it is washed out. Conversely, the WaveNet model amplifies the behavior, adding in new harmonics not present in the original note and modulating all the harmonics. This is seen in the figure by four vertical black stripes that align with the four modulations of the original signal (D). 

\subsection{Entanglement of Pitch and Timbre}
By conditioning on pitch during training, we hypothesize that we should be able to generate multiple pitches from a single $Z$ vector that preserve the identity of timbre and dynamics. Our initial attempts were unsuccessful, as it seems our models had learned to ignore the conditioning variable. We investigate this further with classification and correlation studies.

\begin{figure*}[t]
    \includegraphics[width=\textwidth]{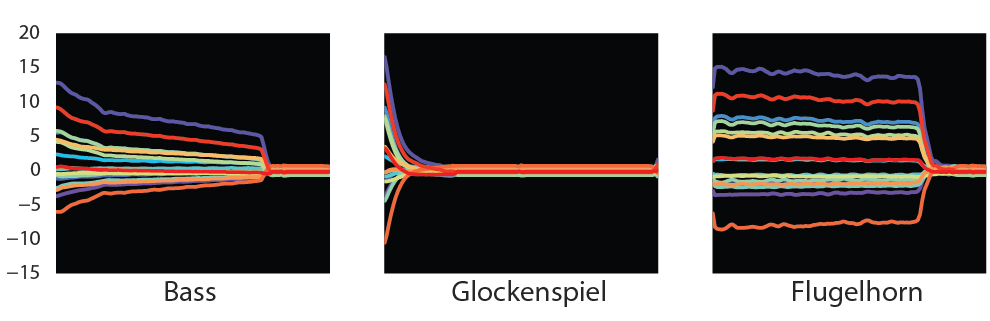}
    \caption{Temporal embeddings for three different instruments. The different colors represent the 16 different dimensions of the embeddings for 125 timesteps (each 32ms). Note that the embedding have a contour similar to the magnitude contour of the original note and decay close to zero when there is no sound. With this in mind, they can be thought of as a "driving function" for a nonlinear oscillator / infinite impulse response filter.}
    \label{fig:Embeddings}
\end{figure*}

\begin{figure*}[t]
    \includegraphics[width=\textwidth]{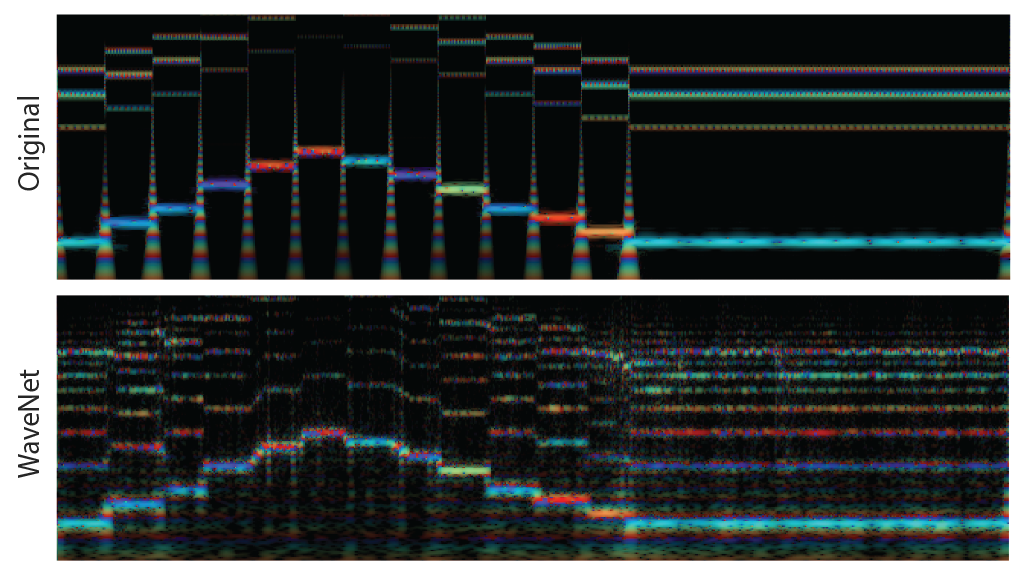}
    \caption{Rainbowgrams of a series of notes reconstructed by the WaveNet autoencoder. The model was never trained on more than one note at a time or on clips longer than four seconds, but it does a fair job of reconstructing this ten-second long scale. (Listen: \href{https://download.magenta.tensorflow.org/audio_examples/nsynth/Figure7_Generalization/Original.mp3}{Original}, \href{https://download.magenta.tensorflow.org/audio_examples/nsynth/Figure7_Generalization/Reconstruction.mp3}{Reconstruction})}
    \label{fig:Extrapolation}
\end{figure*}

\subsubsection{Pitch Classification from $Z$}
\label{sec:PitchShift}

One way to study the entanglement of pitch and $Z$ is to consider the pitch classification accuracy from embeddings. If training with pitch conditioning disentangles the representation of pitch and timbre, then we would expect a linear pitch classifier trained on the embeddings to drop in accuracy. To test this, we train a series of baseline autoencoder models with different embedding sizes, both with and without pitch conditioning. For each model, we then train a logistic regression pitch classifier on its embeddings and test on a random sample of 4096 held-out embeddings.

\begin{table}[t]
\caption{Classification accuracy (in percentage) of a linear pitch classifier trained on learned embeddings. The decoupling of pitch and embedding becomes more pronounced at smaller embedding sizes as shown by the larger relative decrease in classification accuracy.}
\label{table:classify_pitch}
\begin{center}
\begin{small}
\begin{sc}
\begin{tabular}{l c c c c}
\abovespace
 & $Z$ & No Pitch & Pitch & Relative \\
 & Size & Cond. & Cond. & Change\\
\hline\abovespace
WaveNet  & 1984 & 58.1 & 40.5 & \textbf{-30.4}\\
Baseline & 1984 & 63.8 & 55.2 & \textbf{-13.5}\\
Baseline & 1024 & 57.4 & 42.1 & -26.7\\
Baseline & 512 & 63.2 & 21.8 & -65.5\\
Baseline & 256 & 57.7 & 21.0 & -63.6\\
Baseline & 128 & 58.2 & 21.2 & -63.6\\
Baseline & 64  & 59.8 & 15.2 & \textbf{-74.6}\\
\hline
\end{tabular}
\end{sc}
\end{small}
\end{center}
\end{table}

The first two rows of Table~\ref{table:classify_pitch} demonstrate that the baseline and WaveNet models decrease in classification accuracy by 13-30\%  when adding pitch conditioning during training. This is indicative a reduced presence of pitch information in the latent code and thus a decoupling of pitch and timbre information. Further, as the total embedding size decreases below 512, the accuracy drop becomes much more pronounced, reaching a 75\% relative decrease. This is likely due to the greater expressivity of larger embeddings, where there is less to be gained from utilizing the pitch conditioning. However, as the embedding size decreases, so too does reconstruction quality. This is more pronounced for the WaveNet models, which have farther to fall in terms of sample quality.

As a proof of principle, we find that for a baseline model with an embedding size of 128, we are able to successfully balance reconstruction quality and response to conditioning. Figure~\ref{fig:PitchShift} demonstrates two octaves of a C major chord created from a single embedding of an electric piano note, but conditioned on different pitches. The resulting harmonic structure of the original note is only partially preserved across the range. As we shift the pitch upwards, a sub-harmonic emerges (A) such that the pitch +12 note is similar to the original except that the harmonics of the octave are accentuated in amplitude. This aligns with our pitch classification results, where we find that pitches are most commonly confused with those one octave away (see Appendix). These errors can account for as much as 20\% absolute classification error.

\subsubsection{Z Correlation across Pitch}

We can gain further insight into the relationship between timbre and pitch by examining the correlation of WaveNet embeddings among pitches for a given instrument. Figure~\ref{fig:Corr} shows correlations for several instruments across their entire 88 note range at velocity 127. We see that each instrument has a unique partitioning into two or more registers over which notes of different pitches have similar embeddings. Even the average over all instruments shows a broad distinction between high and low registers. On reflection, this is unsurprising as the timbre and dynamics of an instrument can vary dramatically across its range.

\subsection{Generalization of Temporal Encodings}

The WaveNet autoencoder model has some unique properties that allow it to generalize to situations not in the dataset. Since the model learns embeddings that bias an autoregressive decoder, they effectively act as a "driving function" for a nonlinear oscillator / infinite impulse response filter. This is made clear by Figure~\ref{fig:Embeddings}, where the embeddings follow a magnitude contour similar to that of the rainbowgrams of their corresponding sounds in Figures~\ref{fig:Reconstruction} and \ref{fig:InterpolateZ}. 

Further, much like a spectrogram, the embeddings only capture a local context. This lets them generalize in time. The model has only ever seen single notes with sound that lasts for up to three seconds, and yet Figure~\ref{fig:Extrapolation} demonstrates that it can successfully reconstruct both a whole series of notes, as well as notes played for longer than three seconds. While the WaveNet autoencoder adds more harmonics to the original timbre of the organ instrument, it follows the fundamental frequency as it plays up two octaves of a C major arpeggio, back down a G dominant arrpeggio, and holds for several seconds on the base note. The fact that it has never seen a transition between two notes is clear, as the fundamental frequency actually glissandos smoothly between new notes.




\section{Conclusion and Future Directions}
In this paper, we have introduced a WaveNet autoencoder model that captures long term structure without the need for external conditioning and demonstrated its effectiveness on the new NSynth dataset for generative modeling of audio.

The WaveNet autoencoder that we describe is a powerful representation for which there remain multiple avenues of exploration. It builds upon the fine-grained local understanding of the original WaveNet work and provides access to a useful hidden space. However, due to memory constraints, it is unable to fully capture global context. Overcoming this limitation is an important open problem.

NSynth was inspired by image recognition datasets that have been core to recent progress in deep learning. Similar to how many image datasets focus on a single object per example, NSynth hones in on a single note. Indeed, much modern music production employs such a factorization, using MIDI for note sequences and software synthesizers for timbre. Note-to-note dependencies can be partly restored by passing sequence-level timbre and dynamics information to the note-level synthesizer. While not perfect, this factorization is based on the physics of many instruments and is surprisingly effective.

We encourage the broader community to use NSynth as a benchmark and entry point into audio machine learning. We also view NSynth as a building block for future datasets and envision a high-quality multi-note dataset for tasks like generation and transcription that involve learning complex language-like dependencies.

 


\bibliographystyle{icml2013}

\clearpage

\begin{appendices}

\section{Phase Representation for the Baseline Model}
\begin{figure*}[t]
    \includegraphics[width=\textwidth]{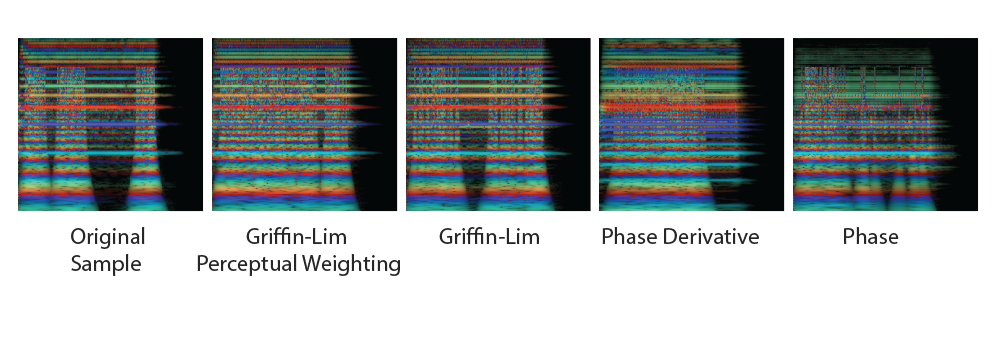}
    \caption{Reconstructions from baseline models trained with different phase representations. For Griffin-Lim, only the magnitude is modeled, and an 1000 iterations of an iterative technique is used to estimate the phase. (Listen: 
    \href{https://download.magenta.tensorflow.org/audio_examples/nsynth/FigureS1_PhaseBaseline/4-Original.mp3}{Original},    
    \href{https://download.magenta.tensorflow.org/audio_examples/nsynth/FigureS1_PhaseBaseline/3-GriffinLimPerceptualWeighting.mp3}{Griffin-Lim Perceptual Weighting},    
    \href{https://download.magenta.tensorflow.org/audio_examples/nsynth/FigureS1_PhaseBaseline/2-GriffinLim.mp3}{Griffin-Lim},    
    \href{https://download.magenta.tensorflow.org/audio_examples/nsynth/FigureS1_PhaseBaseline/1-PhaseDerivative.mp3}{Phase Derivative},    
    \href{https://download.magenta.tensorflow.org/audio_examples/nsynth/FigureS1_PhaseBaseline/0-Phase.mp3}{Phase})  
}
    \label{fig:BaselinePhase}
\end{figure*}

We explored several audio representations for our baseline model. Each representation uses an MSE cost and always includes the magnitude of the STFT spectrogram. We found that training on the peak-normalized log magnitude of the power spectra correlated better with perceptual distortion. When using phase in teh objective, we regress on the phase angle. We can assume a circular normal distribution \cite{Bishop:2006:PRM:1162264} for the phase with a log likelihood loss proportional to $cos(\pi*(x - \hat{x}))$. Figure~\ref{fig:BaselinePhase} shows CQT spectrograms of reconstructions of a trumpet sound from models trained on each input representation. We also include audio of each reconstruction, which is essential listening to hear the improvement of the perceptual weighting.

\section{Description of Quality Tags}
We provide quality annotations for the 10 different note qualities described below. None of the tags are mutually exclusive by definition except for Bright and Dark. However, it is possible for a note to be neither Bright nor Dark.

\begin{itemize}  
\item \textbf{Bright}: A large amount of high frequency content and strong upper harmonics.
\item \textbf{Dark}: A distinct lack of high frequency content, giving a muted and bassy sound. Also sometimes described as 'Warm'.
\item \textbf{Distortion}: Waveshaping that produces a distinctive crunchy sound and presence of many harmonics. Sometimes paired with non-harmonic noise.       
\item \textbf{Fast Decay}: Amplitude envelope of all harmonics decays substantially before the 'note-off' point at 3 seconds.     
\item \textbf{Long Release}: Amplitude envelope decays slowly after the 'note-off' point, sometimes still present at the end of the sample at 4 seconds.
\item \textbf{Multiphonic}: Presence of overtone frequencies related to more than one fundamental frequency.
\item \textbf{Non-Linear Envelope}: Modulation of the sound with a distinct envelope behavior different than the monotonic decrease of the note. Can also include filter envelopes as well as dynamic envelopes.
\item \textbf{Percussive}: A loud non-harmonic sound at note onset.
\item \textbf{Reverb}: Room acoustics that were not able to be removed from the original sample. 
\item \textbf{Tempo-Synced}: Rhythmic modulation of the sound to a fixed tempo. 
\end{itemize}

\begin{table*}[t]
\caption{Instrument annotations. Instruments are labeled with both a source and a family. The source denotes how each instrument's notes are generated: acoustic instrument, electronic instrument, or by software synthesis. The family denotes a high-level class for each instrument.}
\label{instrument-classes}
\vskip 0.1in
\begin{center}
\begin{small}
\begin{sc}
\begin{tabularx}{\columnwidth}{l | X X X | c}
&\multicolumn{3}{|c|}{\textbf{Source}} \\
\textbf{Family} &  Acoust & Electr & Synth & \textbf{Total} \\
\hline\abovespace
Bass       & \num{200}   & \num{8387}  & \num{60368} & \num{68955} \\
Brass      & \num{13760} & \num{70}    & \num{0}     & \num{13830} \\
Flute      & \num{6572}  & \num{70}    & \num{2816}  & \num{9458} \\
Guitar     & \num{13343} & \num{16805} & \num{5275}  & \num{35423} \\
Keyboard   & \num{8508}  & \num{42709} & \num{3838}  & \num{55055} \\
Mallet     & \num{27722} & \num{5581}  & \num{1763}  & \num{35066} \\
Organ      & \num{176}   & \num{36401} & \num{0}     & \num{36577} \\
Reed       & \num{14262} & \num{76}    & \num{528}   & \num{14866} \\
String     & \num{20510} & \num{84}    & \num{0}     & \num{20594} \\
Synth Lead & \num{0}     & \num{0}     & \num{5501}  & \num{5501} \\
Vocal      & \num{3925}  & \num{140}   & \num{6688}  & \num{10753} \\
\hline\abovespace\belowspace
\textbf{Total}      & \num{108978} & \num{110224} & \num{86777} & \num{306043}
\belowspace
\end{tabularx}
\end{sc}
\end{small}
\end{center}
\vskip -0.1in
\end{table*}

\begin{table*}[t]
\caption{Co-occurrence probabilities and marginal frequencies of quality annotations. Both are presented as percentages.}
\label{quality-classes}
\vskip 0.15in
\begin{center}
\begin{small}
\begin{sc}
\begin{tabular}{ c | l | c c c c c c c c c c c}
\abovespace\belowspace
& \textbf{Quality} & \rot{Bright} & \rot{Dark} & \rot{Distortion} & \rot{Fast Decay} & \rot{Long Release} & \rot{Multiphonic} & \rot{Nonlinear Env} & \rot{Percussive} & \rot{Reverb} & \rot{Tempo-Synced} \\
\hline\abovespace
&Dark          & 0.0 \\
&Distortion    & 25.9 & 2.5 \\
&Fast Decay    & 10.0 & 7.5 & 8.1 \\
&Long Release  & 9.0 & 5.2 & 9.8 & 0.0 \\
&Multiphonic   & 6.0 & 1.5 & 5.4 & 2.8 & 6.9 \\
&Nonlinear Env & 8.5 & 1.4 & 6.6 & 2.1 & 6.7 & 8.6 \\
&Percussive    & 6.2 & 5.1 & 3.0 & 52.0 & 0.8 & 2.4 & 0.9 \\
\rot{\rlap{\textbf{Co-occurrence}}}
&Reverb        & 6.6 & 8.9 & 0.3 & 13.0 & 13.7 & 0.7 & 3.5 & 12.4 \\
&Tempo-Synced  & 2.4 & 1.8 & 5.2 & 0.4 & 6.4 & 9.3 & 2.3 & 1.5 & 0.0 \\
\hline\abovespace\belowspace
& \textbf{Frequency} &  13.5 &  11.0 &  17.0 &  14.7 &  8.5 &  3.4 &  3.2 &  10.2 &  16.8 &  1.8
\end{tabular}
\end{sc}
\end{small}
\end{center}
\vskip -0.1in
\end{table*}

 \begin{figure*}[t]
    \includegraphics[width=\textwidth]{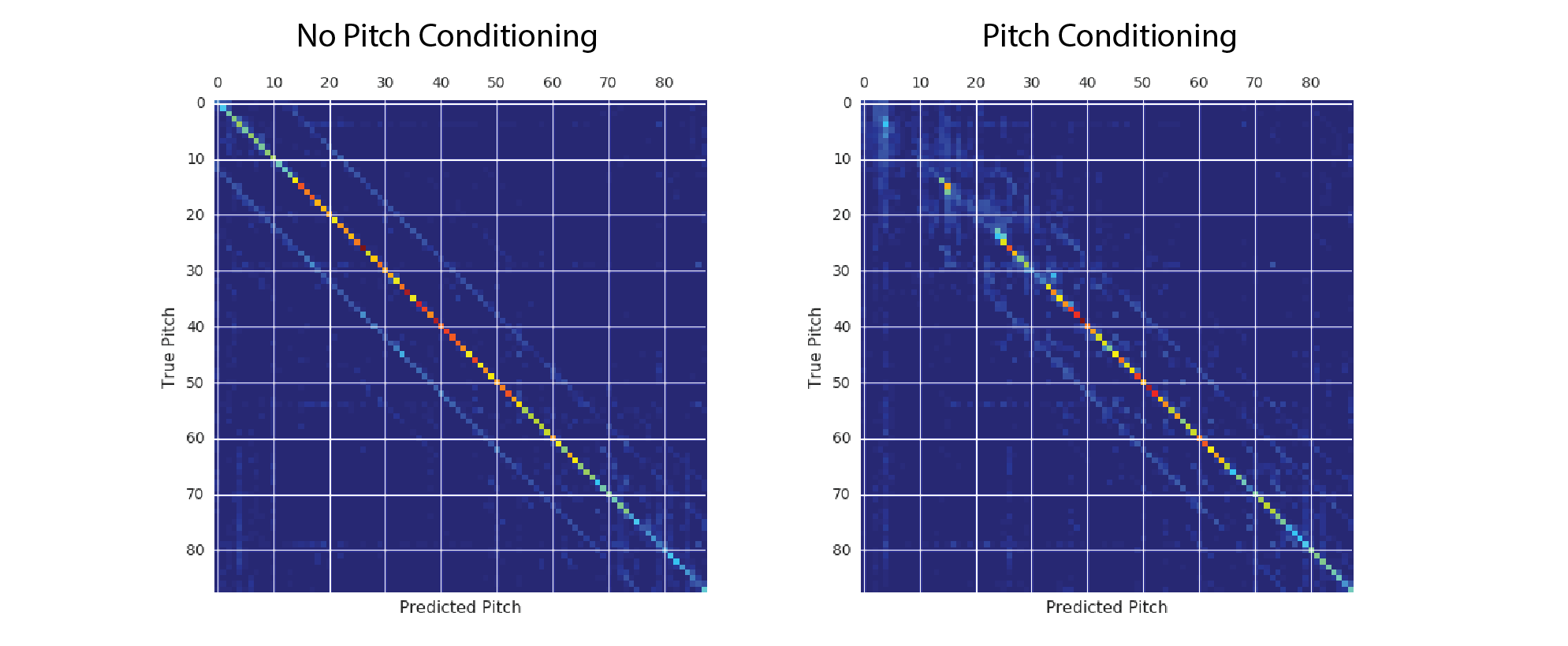}
    \caption{Confusion matrix for linear pitch classification model trained on embeddings from a WaveNet autoencoder. The prodominant error is predicting the wrong octave (being off by 12 tones). Training with pitch conditioning reduces the classifer accuracy.}
    \label{fig:Confusion}
\end{figure*}

\section{Details of Pitch and Quality Classifier}

\begin{figure*}[t]
\includegraphics[width=\textwidth]{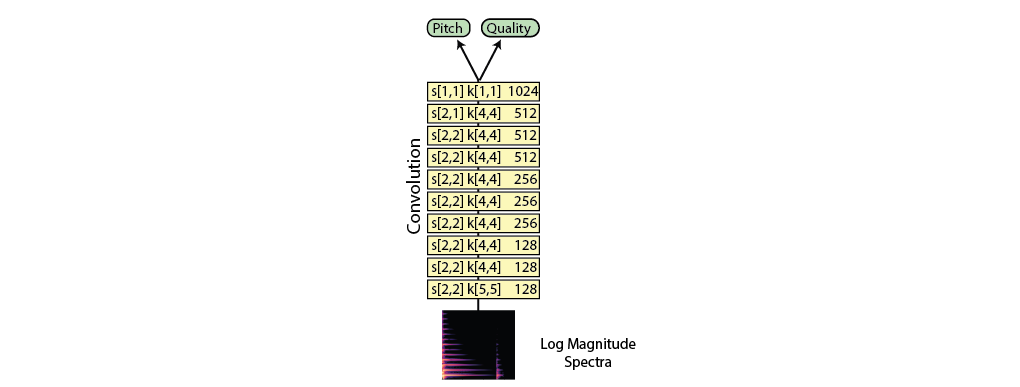}
\caption{Model architecture for pitch and quality classification. Like the baseline encoder, each convolution layer is followed by batch normalization and a Leaky-ReLU (0.1 off-slope). }
\label{Fig:model}
\end{figure*}

We train a multi-task classification model to do pitch and quality tag classification on the entire NSynth dataset. We use the the encoder structure from the baseline model with the exception that there is no bottleneck (see Figure~\ref{Fig:model}). We use a softmax-crossentropy loss for the pitch labels as they are mutually exclusive and a sigmoid-crossentropy loss for the quality tags as they are not. Note that since the architecture uses only magnitude spectra, it cannot take advantage of the improved phase coherence of the WaveNet samples.

\end{appendices}

\end{document}